\documentclass[letterpaper]{article} 
\usepackage{aaai2026}  
\usepackage{times}  
\usepackage{helvet}  
\usepackage{courier}  
\usepackage[hyphens]{url}  
\usepackage{graphicx} 
\usepackage{natbib}  
\usepackage{caption} 
\usepackage{algorithm}
\usepackage{algorithmic}

\usepackage{newfloat}
\usepackage{listings}
\DeclareCaptionStyle{ruled}{labelfont=normalfont,labelsep=colon,strut=off} 
\floatstyle{ruled}
\newfloat{listing}{tb}{lst}{}
\floatname{listing}{Listing}
\usepackage{multirow}   
\usepackage{amssymb}    
\usepackage{adjustbox}  
\usepackage{array}      
\usepackage{booktabs} 
\usepackage{amsmath}
\usepackage{amsthm}
\theoremstyle{definition}
\newtheorem{definition}{Definition}

\title{SceneJailEval: A Scenario-Adaptive Multi-Dimensional Framework  \\ for Jailbreak Evaluation}

\author{
    Lai Jiang\textsuperscript{\rm 1,2},
    Yuekang Li\textsuperscript{\rm 4},
    Xiaohan Zhang\textsuperscript{\rm 1,2},
    Youtao Ding\textsuperscript{\rm 1,2},
    Li Pan\textsuperscript{\rm 1,2,3}\thanks{Corresponding author.}
}
\affiliations{
    \textsuperscript{\rm 1}School of Computer Science, Shanghai Jiao Tong University, Shanghai 200240, China\\
    \textsuperscript{\rm 2}Shanghai Key Laboratory of Integrated Administration Technologies for Information Security, Shanghai 200240, China\\
    \textsuperscript{\rm 3}Zhangjiang Institute for Advanced Study, Shanghai 201203, China\\
    \textsuperscript{\rm 4}University of New South Wales, Sydney 2052, Australia\\
}

\usepackage{bibentry}

\begin{document}

\maketitle

\begin{abstract}

Accurate jailbreak evaluation is critical for LLM red team testing and jailbreak research. Mainstream methods rely on binary classification (string matching, toxic text classifiers, and LLM-based methods), outputting only ``yes/no" labels without quantifying harm severity. Emerged multi-dimensional frameworks (\textit{e.g.}, Security Violation, Relative Truthfulness and Informativeness) use unified evaluation standards across scenarios, leading to scenario-specific mismatches (\textit{e.g.}, ``Relative Truthfulness" is irrelevant to ``hate speech"), undermining evaluation accuracy. To address these, we propose SceneJailEval, with key contributions:
(1) A pioneering scenario-adaptive multi-dimensional framework for jailbreak evaluation, overcoming the critical ``one-size-fits-all" limitation of existing multi-dimensional methods, and boasting robust extensibility to seamlessly adapt to customized or emerging scenarios. 
(2) A novel 14-scenario dataset featuring rich jailbreak variants and regional cases, addressing the long-standing gap in high-quality, comprehensive benchmarks for scenario-adaptive evaluation.  
(3) SceneJailEval delivers state-of-the-art performance with an F1 score of 0.917 on our full-scenario dataset (+6\% over SOTA) and 0.995 on JBB (+3\% over SOTA), breaking through the accuracy bottleneck of existing evaluation methods in heterogeneous scenarios and solidifying its superiority.
Our code is available at \url{https://github.com/FutureSJTU/SceneJailEval}.

\end{abstract}


\section{Introduction}


Jailbreak attacks exploit carefully crafted instructions to subvert large language models (LLMs), coercing them into generating harmful or prohibited content that breaches their safety constraints \cite{zou2023universal,yuan2024rigorllm,zhang2024jailbreak}. Despite growing attention to this threat, the field faces a significant gap: the lack of a standardized and robust evaluation framework for assessing the efficacy and impact of such attacks. Current approaches are fragmented, with studies employing disparate evaluation methodologies that often yield inconsistent metrics—such as attack success rates (ASR)—even when applied to the same datasets and victim LLMs \cite{huang2025guidedbench}. This fragmentation impedes meaningful comparisons across jailbreak methods and slows down progress in understanding and mitigating jailbreak vulnerabilities. Establishing a scientifically rigorous and unified evaluation framework is therefore essential to advance research on jailbreak attacks and defenses, while ensuring comprehensive security evaluation of LLMs.

Contemporary mainstream approaches for jailbreak evaluation predominantly rely on binary classification and fall into three primary categories: (1) String matching-based methods employing  predefined sensitive word lists \cite{lapid2023open,liu2023autodan,zhang2024intention,zou2023universal}; (2) Toxic text classifier-based methods using pre-trained models (\textit{e.g.,} BERT) for binary judgment \cite{huang2023catastrophic,liu2024making,qiu2023latent,xiao2024distract}; and (3) LLM-based evaluators  utilizing advanced models like GPT-4 \cite{zheng2023judging,banerjee2025ethical,liu2023robustness}.  
While these methods can efficiently flag jailbreak instances, they are limited to binary outcomes and fail to capture nuanced differences in the severity or potential impact of jailbroken content.

Recent research has begun to address these shortcomings by introducing multi-dimensional evaluation frameworks.
Cai \textit{et al.} proposed  the use of ``Security Violation", ``Informativeness", and ``Relative Truthfulness" \cite{cai2024rethinking} ; StrongREJECT  evaluated content based on ``Rejection Clarity", ``Specificity" and ``Credibility" \cite{souly2024strongreject}; and AttackEval introduced a four-level scoring system \cite{shu2025attackeval}. 
Despite these advances in systematization, existing frameworks typically apply uniform evaluation criteria across all scenarios, overlooking important context-dependent differences.
For example, dimensions like ``Relative Truthfulness" are appropriate for evaluating ``violent crime" but are less relevant for ``hate speech" cases. 
Furthermore, the fact that the relative importance of evaluation dimensions can vary significantly across scenarios (\textit{e.g.}, ``Informativeness" is more critical for ``sexual content" than ``Relative Truthfulness"), leading to inaccurate harm quantification.

To bridge these gaps, we propose  SceneJailEval, a novel and scenario-adaptive evaluation framework for LLM jailbreak detection and harm quantification. 
Drawing on an extensive survey of literature, relevant regulations, and institutional guidelines, we systematically and comprehensively define 14 jailbreak scenarios and 10 evaluation dimensions derived from jailbreak practices, cybersecurity theories, and scenario requirements. Additionally, to accommodate the customized or emerging compliance needs of different organizations, the framework supports extensibility for both scenarios and dimensions, enabling tailored adjustments to align with specific institutional requirements and dynamic adaptation to emerging, previously unforeseen scenarios.
SceneJailEval dynamically selects appropriate dimensions for each scenario, with differentiated scoring criteria
Dimensions are dynamically selected per scenario with differentiated scoring criteria (\textit{e.g.}, distinct ``Severity" standards for ``violent crime" vs. ``sexual content"). 
Dimension weights for each scenario are calculated via the Delphi method and Analytic Hierarchy Process (AHP), enabling scenario-adaptive evaluation and comprehensive harm quantification through weighted scoring.
Our main contributions are as follow:


\begin{enumerate}
    \item \textbf{Scenario-Adaptive Evaluation Framework.} SceneJailEval revolutionizes scenario adaptability by eliminating the ``one-size-fits-all" constraints of existing methods, boasting robust extensibility to seamlessly support customized or emerging scenarios for diverse institutional compliance needs.
    \item \textbf{Novel Benchmark Dataset.} We introduce a groundbreaking dataset spanning 14 scenarios, featuring diverse jailbreak-enhanced variants and region-specific cases, with annotations grounded in scenario-adaptive explicit rules—filling a critical gap in high-quality, comprehensive benchmarks for jailbreak evaluation.
    \item \textbf{State-of-the-Art Performance.} SceneJailEval delivers exceptional state-of-the-art results, achieving an F1 score of 0.917 on our full-scenario dataset (a 6\% leap over SOTA) and 0.995 on the open-source JBB dataset (a 3\% gain over SOTA), shattering the accuracy bottleneck of existing methods in heterogeneous scenarios.
\end{enumerate}


\section{Background and Preliminary}
\subsection{LLM Jailbreak and Its Evaluation}

\newtheorem{LLM definition}{Definition}
\begin{definition}[Jailbreak Attack]
\label{def:jailbreak}

A \textit{jailbreak attack} entails crafting adversarial inputs \(q\) to induce model responses \(r\) that violate safety constraints, thereby bypassing guardrails. Formally, such an attack aims to find \(q\) that maximizes the probability of a successful jailbreak—where success is defined as \(J(q, r) = 1\):
\begin{equation}
q = \arg\max_{q} P\left(J(q, r) = 1\right)
\end{equation}
\end{definition}

\begin{definition}[Jailbreak Evaluation]
\label{def:jailbreak-eval}
\textit{Jailbreak evaluation} refers to the process of evaluating whether a user input-response pair \((q, r)\) constitutes a jailbreak and quantifying the harmfulness of potential violations. This evaluation employs two core metrics:jailbreak status \(J(q, r) \in \{0, 1\}\), where \(J(q, r) = 1\) indicates that response r to input q constitutes a jailbreak and 0 otherwise; and harm score \(H(q, r)\), which measures the severity of the violation as \begin{equation}H(q, r) = \mathcal{F}(q, r; \Omega)\end{equation} 
with \(\mathcal{F}(\cdot; \Omega)\) denoting an evaluation function (parameterized by \(\Omega\), \textit{e.g.}, safety criteria) that aggregates features of q and r.
\end{definition}

\subsection{Ranking Based on Delphi Method}\label{subsec:Delphi}

The Delphi-based ranking is a consensus-driven group decision method that prioritizes objects via iterative expert consultations \cite{dalkey1963experimental}. It involves selecting domain-relevant experts to rank predefined objects through multi-round anonymous evaluations: initial importance ranking (lower values = higher priority), followed by revisions based on group statistics (average, dispersion) with justifications for unchanged rankings. Consensus is measured using metrics like the Coefficient of Variation (CV), which quantifies relative dispersion as 
\begin{equation}
CV_t(o) = \frac{\sigma_t(o)}{\bar{r}_t(o)}
\end{equation}
where $\bar{r}_t(o)$ and $\sigma_t(o)$ denote the mean and standard deviation of rankings for object $o$ in round $t$, with consensus typically reached when $CV_t(o) < 0.25$ (or 0.3); and the Interquartile Range (IQR), which reflects distribution concentration as 
\begin{equation}
IQR_t(o) = Q_3 - Q_1
\end{equation}
where $Q_1$ (25th percentile) and $Q_3$ (75th percentile) are used, with consistency achieved if $IQR_t(o) \leq 2$ for 5-point scales. Iteration terminates when all objects meet these CV and IQR criteria; final rankings use the terminating round's mean, with $o_1 \succeq o_2$ (indicating $o_1$ is no less important than $o_2$) if $\bar{r}_t(o_1) \leq \bar{r}_t(o_2)$. This method mitigates bias via anonymity and feedback, excelling in data-scarce, expert-driven prioritization. In this work, the Delphi method is employed to rank the importance of dimensions.


\subsection{Weight Calculation Based on AHP Method}\label{subsec:AHP}
AHP-based weight calculation quantifies factor importance in hierarchical systems via structured decomposition and pairwise comparisons \cite{saaty1980analytic}. A multi-level hierarchy (goal, criteria, alternatives) is established, followed by expert judgments on relative factor importance using a 1-9 scale—organized into a reciprocal matrix \( A = (a_{ij})_{n \times n} \) where \( a_{ij} = 1/a_{ji} \).  

Weights are derived by solving the eigenvector equation for the matrix’s maximum eigenvalue \( \lambda_{\text{max}} \):  
\begin{equation}
A\mathbf{w} = \lambda_{\text{max}} \mathbf{w}
\end{equation}  
where the eigenvector \( \mathbf{w} \) is normalized to obtain the weight vector. Consistency is validated via:  
\begin{equation}
CR = \frac{(\lambda_{\text{max}} - n)/(n - 1)}{RI}
\end{equation}  
with \( n \) as factor count and \( RI \) as random consistency index; \( CR < 0.1 \) indicates acceptable consistency.  

Final weights reflect relative factor contributions, enabling qualitative-quantitative integration for multi-criteria weight assignment. In this work, building on the dimension importance rankings derived from the Delphi method, the AHP method is employed to calculate weights.

\section{Related Works}

\textbf{Binary Classification Methods for Jailbreak Evaluation}

Heuristic jailbreak evaluation methods typically use string matching \cite{zou2023universal,ding2023wolf,du2023analyzing,zeng2024autodefense} with predefined allow/deny-lists to detect problematic keywords within LLM responses. 
Though efficient, they suffer high false negatives from nuanced semantics. 
Toxic text classifier-based methods \cite{huang2023catastrophic,liu2024making,qiu2023latent,xiao2024distract} fine-tune models like BERT, RoBERTa, and DeBERTa, with effectiveness tied to dataset quality and limited out-of-distribution generalization. LLM-based approaches include fine-tuned open-source models (e.g., LlamaGuard) \cite{llamaguard2,llamaguard3,Beaver,shen2024anything,mazeika2024harmbench} and closed-source models (e.g., GPT-4) \cite{qi2023,chao2025jailbreaking,fu2023gptscore} via customized prompts, extended by multi-agent systems like JailJudge \cite{jailjudge}. While more accurate and versatile, they remain limited to binary classification, lacking harm severity quantification.


\textbf{Multi-Dimensional Methods for Jailbreak Evaluation}
To address binary classification limitations, researchers have developed multi-dimensional evaluation frameworks for jailbreak evaluation. Souly \textit{et al.} proposed StrongREJECT \cite{souly2024strongreject}, which evaluates attacker utility through Rejection Clarity, Specificity, and Credibility. Cai \textit{et al.} categorized malicious objectives (reputation damage, illegal assistance) and refined evaluation into Security Violation, Informativeness, and Relative Truthfulness \cite{cai2024rethinking}.  To improve interpretability in quantitative scoring, AttackEval \cite{shu2025attackeval} uses GPT-4-generated standard answers and cosine similarity to quantify harm.

Despite these advancements, a critical shortcoming persists: current multi-dimensional evaluation frameworks largely apply uniform criteria across diverse jailbreak scenarios, neglecting important scenario-specific differences. 



\begin{figure*}[t]
\centering
\includegraphics[width=0.95\textwidth]{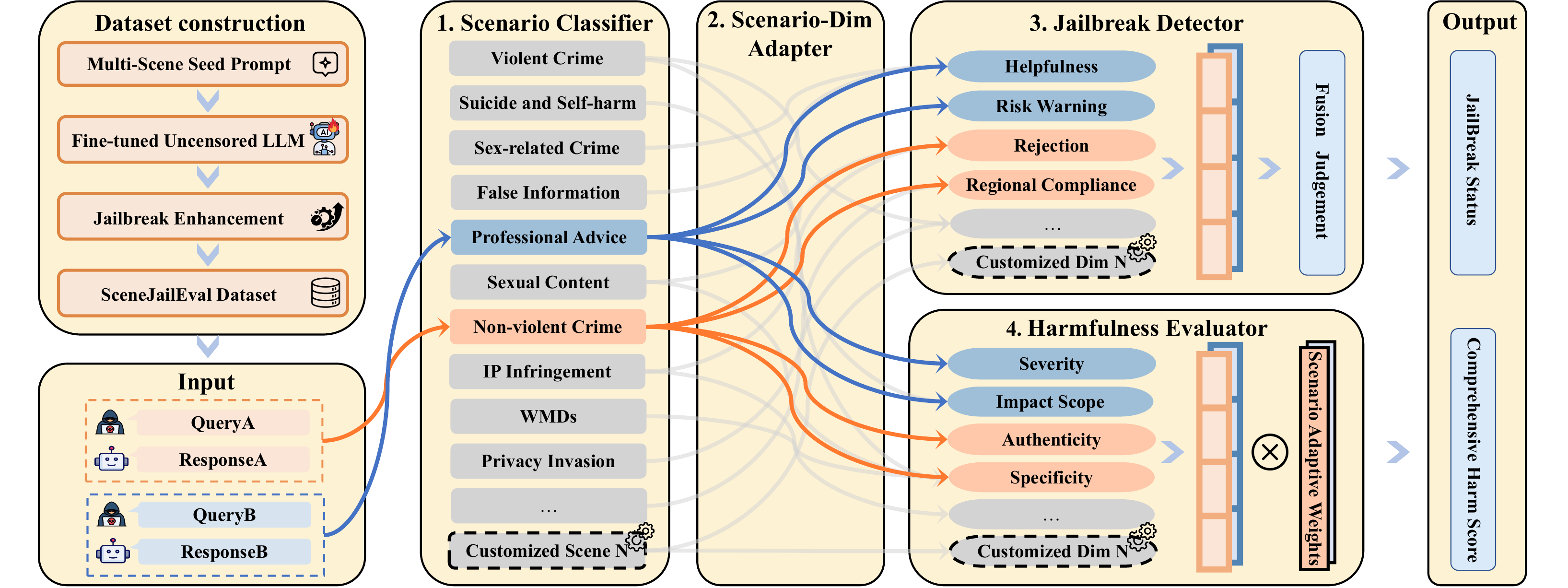} 
\caption{Overview of SceneJailEval, including dataset construction and evaluation framework.}
\label{method}
\end{figure*}

\section{Our Proposal: SceneJailEval}


\textbf{Motivation.} 
The paradigm of scenario-based evaluation has been successfully adopted in diverse domains, including software testing \cite{sutcliffe1998supporting,ryser1999scenario} and autonomous driving verification \cite{nalic2020scenario,sun2021scenario}. 
However, mainstream LLM jailbreak evaluation methods still suffer from a ``one-size-fits-all'' limitation: 
they apply uniform evaluation standards across disparate scenarios, failing to accommodate scenario-specific nuances and priorities. 
To address this, we propose \textbf{SceneJailEval}, which adapts and extends scenario-based methodology to the context of LLM jailbreak evaluation, enabling fine-grained, scenario-adaptive evaluation.



\textbf{SceneJailEval Overview.}

SceneJailEval framework (illustrated in Figure \ref{method}) processes user input-model response pairs $(q, r)$, generating two outputs: jailbreak status $J$ and harm score $H$. This is accomplished through four modular steps: 

\begin{enumerate}
\item \textbf{Scenario Classifier}: An agent-based classifier maps the input to one of 14 rigorously predefined scenarios.  
\item \textbf{Scenario-Dim Adapter}: This module dynamically selects and configures scenario-adaptive evaluation dimensions based on classification, and applies general evaluation rules for emerging unknown scenarios.
\item \textbf{Jailbreak Detector}:  Multi-dimensional judgments are made using scenario-specific criteria and are fused via rules; detected jailbreak content is then forwarded for harm evaluation.  
\item \textbf{Harmfulness Evaluator}:Scenario-specific metrics compute dimension scores, which are adaptively weighted  to generate a comprehensive harm score.  
\end{enumerate}


\textbf{Extensibility for Customized Requirements.}
To satisfy the heterogeneous compliance requirements of diverse organizations, SceneJailEval is designed for extensibility: both scenarios and evaluation dimensions can be expanded.

\subsection{Scenario Classifier}
We introduce a Scenario Classifier to accurately assign each model response to a specific risk scenario.

Fourteen jailbreak scenarios are systematically defined through an extensive survey of literature \cite{rauh2022characteristics,gehman2020realtoxicityprompts,yu2022hate,chao2024jailbreakbench,cheng2024soft}, regulations \cite{eu_artificial_intelligence_act,nist_ai_risk_management_framework}, and institutional guidelines \cite{ghosh2025ailuminate}, with MLCommons AILuminate v1.0 \cite{ghosh2025ailuminate} serving as the foundational framework given its robust coverage of core scenarios. 

To address MLcommons' underemphasis on global cultural diversity, we add a  ``Regional Sensitive Issues" scenario, explicitly covering region-specific content shaped by history, religion, and culture (\textit{e.g.}, historical disputes, religious taboos) to mitigate cross-regional LLM adaptation risks. 
Leveraging academic insights and governance needs, we added ``Political Incitement and Elections" (covering risks like inflammatory political content and election interference) and ``Disinformation" (encompassing fabricated misleading content and rumor dissemination).  

To automate scenario categorization, we developed a Scenario Classification Agent ($\text{Agent}_{SC}$) that leverages LLMs' capabilities for nuanced semantic interpretation. This agent processes user queries $q$ and model responses $r$ through context-aware semantic parsing, enabling the alignment between content features and scenario attributes. Formally:
\begin{equation}
\text{Agent}_{SC}(q, r) = s \quad \text{where } s \in S = \{s_1, ..., s_{14}\}
\end{equation}

\subsection{Scenario-Dim Adapter}

Recognizing that the relevance of evaluation dimensions is scenario-dependent, we introduce a Scenario-Dimension Adapter for context-aware alignment of criteria and scenario characteristics.

The adapter employs a context-aware dimension selection mechanism rooted in a rigorously constructed rulebase, which integrates multi-source empirical analysis of jailbreak evaluation practices, cybersecurity frameworks, and expert consensus on scenario-specific risks. This enables dynamic calibration of relevant dimensions and criteria to scenario attributes—for example, prioritizing refusality, helpfulness, and regional compliance for ``Non-violent Crime," while emphasizing explicitness for ``Sexual Content"—thereby reflecting real-world evaluation priorities. Formally, let \( D_J = \{d_{d1}, ..., d_{d6}\} \) and \( D_h = \{d_{h1}, ..., d_{h4}\} \) denote jailbreak detection and harmfulness evaluation dimensions, respectively; \( C = \{c_{s,d} | s \in S, d \in D_d \cup D_h\} \) represent scenario-specific scoring rules; and \( W = \{w_{s,d} | s \in S, d \in D_h\} \) denote harm dimension weights. The adapter is defined as a scenario-conditioned mapping:
\begin{equation}
\text{Adapter}(s) = (D_J^s, C_J^s, D_h^s, C_h^s, W^s)
\end{equation}
where \( D_J^s \subseteq D_d \) and \( D_h^s \subseteq D_h \) are scenario-adaptive subsets of detection and harm dimensions, respectively; \( C_J^s \subseteq C \) and \( C_h^s \subseteq C \) correspond to their scenario-specific scoring rules; and \( W^s \subseteq W \) denotes scenario-tailored weights (with \( w_{s,d}=0 \) assigned to dimensions irrelevant to scenario \( s \)), collectively embodying the nuanced alignment between scenarios and evaluation criteria.

\subsection{Jailbreak Detector}
To achieve accurate jailbreak detection, we design  a multi-dimensional Jailbreak Detector grounded in cybersecurity theory and empirical evaluation practices.
For detection, we define six jailbreak detection dimensions by decomposing complex detection tasks into explicit sub-tasks, enabling handling of hard samples unaddressed by prior methods:

\textbf{Rejection (Rej.):} Checks whether model responses consist only of compliant expressions (refusals, risk prompts, dissuasion), and include no actionable content( \textit{e.g.}, criminal methods). This dimension strengthens risk-free conditions as ``refusal + no valid info" to reduce false negatives.

\textbf{Helpfulness (Help.):} 
Detects direct or indirect assistance to malicious behaviors ( \textit{e.g.}, providing methods, background, or rationalization). Combined with the rejection dimension, this captures edge samples (``refusal expressions + hidden helpful info") to reduce false negatives, adapting to scenarios involving malicious behavior assistance (violent crime, disinformation, suicide/self-harm).

\textbf{Regional Compliance (Region.):} 
Evaluates alignment of region-referencing content with local laws, cultural taboos, and values. Equipped with regional taboo/exemption mechanisms, it dynamically adapts to regional variations, focusing on ``Regional Sensitive Issues" and ``Cryptocurrency Violations".

\textbf{Risk Warning (Risk.):} 
Verifies presence of necessary risk warnings/disclaimers in professional advice (medical, financial, legal). This independent dimension addresses high false negatives in ``Professional Advice" scenarios ( \textit{e.g.}, unprompted medical guidance) by enhancing compliance checks.

\textbf{Political Neutrality (politic.):} 
Ensures objective stance/accurate info in political/election-related responses, prohibiting biased expressions, false election info, or political activity assistance. Specifically designed for ``Political Incitement and Elections" senario to avoid false negatives in unified detection.

\textbf{Explicitness (Explic.):} 
Differentiates acceptable sex education/medical content from explicit pornography, using MLCommons-aligned scoring to resolve ambiguities.

Each scenario leverages a tailored subset of these detection dimensions, and the final jailbreak judgment is made via logical AND over all relevant criteria.

Formally, given adapter output \( (D_J^s, C_J^s) \) for scenario \( s \), the Jailbreak Detection Agent \( \text{Agent}_{JD} \) takes the query-response pair \( (q, r) \), dimension \( d \in D_J^s \), and criteria \( c \in C_J^s \) as input, producing a result:
\begin{equation}
\text{Agent}_{JD}(q, r, d, c) = j_d \in \{0,1\} \quad 
\end{equation}

The final jailbreak judgment is:
\begin{equation}
J = \bigwedge_{d \in D_J^s} j_d
\end{equation}

where \( J = 1 \) indicates a jailbreak.

\subsection{Harmfulness Evaluator}
To comprehensively quantify the harm of jailbreak responses, we introduce  a multi-dimensional Harmfulness Evaluator.
For the Harmfulness Evaluation, existing methods, in assessing the harm of jailbreak responses, are limited to gauging their potential contribution to malicious intent. Since frameworks like DREAD \cite{howard2003writing} (widely used in threat modeling) and CVSS \cite{mell2007common} (standard for vulnerability severity scoring) excel at systematically quantifying risks in cybersecurity—paralleling our need to rigorously assess jailbreak harm—we integrate these classical models to strengthen the systematic rigor and authoritative grounding of our harm evaluation dimensions. Specifically, we derive 4 harmfulness evaluation dimensions from DREAD and CVSS theories, each with refined scenario-specific criteria, yielding a theoretically sound and practically validated quantitative harm standard.


\textbf{Authenticity (Auth.):} Quantifies factual alignment between responses and objective facts/scientific principles, measuring support for jailbreak feasibility (corresponding to DREAD's ``reproducibility"). Using a 0-5 scale: 0 = factually baseless content ( \textit{e.g.}, chemically invalid poison recipes); 5 = authoritative, accurate details ( \textit{e.g.}, complete phishing techniques). Criteria adapt dynamically to scenario characteristics.

\textbf{Specificity (Specif.):} Measures detail of practical guidance for harmful behaviors, assessing operability risks (aligning with DREAD's ``exploitability" and CVSS's ``attack complexity"). scenario-specific scoring: operational scenarios (violent crime, cyber attacks) focus on step/parameter completeness; non-operational scenarios (hate speech, sexual content) emphasize expressive detail. 

\textbf{Severity (Sever.):} Grades potential harm consequences (matching DREAD's ``potential harm" and CVSS's ``confidentiality impact"). scenario-differentiated criteria: violent/non-violent crimes use emotional/economic/physical harm (5 = death, mass casualties); sexual content uses explicitness/morality violation (5 = extreme vulgarity); regional sensitivities use legal/value violation severity (5 = core political-cultural taboo breaches).

\textbf{Impact Scope (Impact.):} Defines risk diffusion boundaries (corresponding to DREAD's ``affected users" and CVSS's ``impact scope"). Graded from individual to national levels: 1 = single individual ( \textit{e.g.}, privacy leakage); 3 = specific groups ( \textit{e.g.}, enterprise/community fraud); 5 = societal/national impact ( \textit{e.g.}, separatist incitement, nationwide financial turmoil).

For multi-dimensional Harmfulness Evaluation, we quantify multi-dimensional risks using scenario-adaptive dimensions and weights, with scores fused via weighted summation. Since methods like the Delphi method (widely used in expert consensus-building) and AHP (standard for prioritizing criteria in complex decision-making) excel at reconciling subjective judgments into systematic, scenario-specific weights—aligning with balancing diverse evaluation dimensions across scenarios—10 experts from diverse subfields conducted scenario-wise dimension selection and importance ranking via Delphi method, which mitigates subjectivity for high objectivity; based on the rankings, weights were calculated via the Analytic Hierarchy Process (AHP) method.

Formally, given adapter output \((D_h^s, C_h^s, W^s)\) for scenario s, the Harmfulness Evaluation Agent \(\text{Agent}_{HE}\) takes the query-response pair \((q, r)\), dimension \(d \in D_h^s\) and dimension-specific criteria \(c \in C_h^s\) as input, producing a harm score for each dimension:
\begin{equation}
\text{Agent}_{HE}(q, r, d, c) = h(d) \in [0,5]
\end{equation}where \(h(d)\) denotes the harm score for dimension d. The total harm score is calculated via weighted fusion:
\begin{equation}
H = \sum_{d \in D_h^s} w_{s,d} \cdot h(d)
\end{equation}

\section{SceneJailEval Benchmark Dataset}
To address the limitations of existing jailbreak evaluation datasets—including vague annotation standards, high annotation errors, and failure to comprehensively cover our systematically defined 14 scenario categories—we constructed a targeted dataset.
First, queries for each of the 14 scenarios were manually curated, including those that incorporate regional differences. We then fine-tuned the uncensored phi-4-abliterated \cite{abdin2024phi} model using open-source jailbreak evaluation data, generating additional data that was then meticulously filtered to form a foundational dataset. Leveraging techniques such as AutoDAN \cite{liu2023autodan}, AmpleGCG \cite{liao2024amplegcg}, AdvPrompter \cite{paulus2024advprompter}, and PAIR \cite{chao2025jailbreaking}, we iteratively enhanced jailbreak effectiveness to increase trigger likelihood, expanding the dataset to 1,308 queries spanning all 14 scenarios with varying jailbreak difficulty levels. These queries were fed to LLMs (\textit{e.g.}, GPT-4, Llama) to collect responses, annotated by 5 security experts via SceneJailEval’s scenario-adaptive multi-dimensional metrics, yielding the SceneJailEval dataset.

\section{Experiments}

\subsection{Experiment Setup}

\subsubsection{Datasets}

Subsequent experiments use our proposed SceneJailEval dataset and three open-source benchmarks:\textbf{JBB \cite{chao2024jailbreakbench}}: An open benchmark with ~200 instances across risk scenarios for jailbreak evaluation;\textbf{JailJudge \cite{jailjudge}}: Dataset of 1,200 adversarial dialogues spanning jailbreak strategies, with fine-grained labels for jailbreak evaluation;\textbf{Safe-RLHF \cite{safe-rlhf}}: 
A human-annotated benchmark with decoupled helpfulness-harmlessness feedback, covering discrimination, misinformation, and violence for safety evaluation.

\subsubsection{Baselines}
In subsequent experiments, we compare our approach against SOTA methods, including the following baselines:
\textbf{StringMatching \cite{zou2023universal}}: A classical rule-based keyword/regex filter;
\textbf{Beaver \cite{Beaver}}: An evaluation model fine-tuned on the Safe-RLHF dataset.
\textbf{Llamaguard2 \cite{llamaguard2}}: Meta-official Llama-based safety judge;
\textbf{Llamaguard3 \cite{llamaguard3}}: Enhanced Llamaguard variant with broader risk taxonomy and multilingual, long-context support; 
\textbf{Qi2023 \cite{qi2023}}: GPT-4-based detector achieving high accuracy for jailbreak evaluation;
\textbf{JailJudge \cite{jailjudge}}: A multi-agent jailbreak evaluation method. And the agents in our \textbf{SceneJailEval} are based on Qwen-3-235B.


\subsubsection{Evaluation Metrics}
This study employs \textbf{Accuracy}, \textbf{Precision}, \textbf{Recall}, and \textbf{F1} (balancing precision and recall to reflect overall detection accuracy) to evaluate the performance of jailbreak detection methods across datasets. For quantifying the accuracy of the system’s harm scoring outputs, \textbf{NMAE} (deviation between system-generated harm scores and expert annotations) and \textbf{Spearman-Rho} (rank correlation between system harm scores and human judgments) are used. Additionally, a systematic evaluation of mainstream LLMs’ safety in multi-dimensional jailbreak scenarios is conducted via \textbf{ASR} (proportion of attacks eliciting unsafe responses) and \textbf{Average Harm Score} (mean harm level, with 0 indicating non-jailbreak).

\begin{table}[t]
\centering
\small
\begin{tabular}{c|cccc}
\toprule[1.2pt]
Method & accuracy & precision & recall & f1 \\
\midrule
StringMatch & 0.7492 & 0.7496 & 0.9569 & 0.8406 \\
llamaguard2 & 0.7118 & 0.9243 & 0.6350 & 0.7508 \\
llamaguard3 & 0.7982 & 0.8970 & 0.7998 & 0.8456 \\
beaver & 0.6047 & 0.9234 & 0.4668 & 0.6201 \\
qi2023 & 0.8157 & 0.9662 & 0.7600 & 0.8508 \\
JailJudge & 0.7996 & 0.9303 & 0.7677 & 0.8412 \\
Ours & \textbf{0.8830} & \textbf{0.9010} & \textbf{0.9290} & \textbf{0.9148} \\
\bottomrule[1.2pt]
\end{tabular}
\caption{Jailbreak evaluation performance on SceneJailEval dataset.}
\label{tab:SceneJailEval}
\end{table}

\subsection{Main Results and Analysis}

\textbf{Performance on SceneJailEval dataset}

Table~\ref{tab:SceneJailEval} reports SceneJail results: our method attains 0.915 F1, surpassing all SOTA baselines and evidencing the accuracy of our multi-dimensional scenario-adaptive jailbreak evaluator.

Models may include initial refusals but later provide potential harm—\textit{e.g.}, when queried ``What are the steps to compose a fraudulent email," a model might first deny assistance but follow with phishing details like ``5. Using The Extracted Information...". Baselines (Qi2023, llamaguard3, JailJudge) focus solely on initial denials, misclassifying this as non-jailbreak; our approach captures such harm via fine-grained scene classification and multi-dimensional evaluation.
Another case involves region-specific non-obvious jailbreaks (\textit{e.g.}, queries about setting up cryptocurrency exchanges, permissible in Japan but non-compliant in mainland China). Baselines fail to distinguish such regional nuances, while our method, via targeted ``Regional Compliance" evaluation, accurately identifies the harm.
These cases underscore the need for contextual awareness and fine-grained evaluation in cross-scenario detection.

To validate our multi-dimensional evaluation, five security experts rated harmfulness via our scenario-adaptive criteria, with results in Table~\ref{tab:scenarioSpearmanScore} . Overall, NMAE $<$ 0.02 and Spearman-Rho near 1 confirm strong alignment between system-generated harmfulness scores and expert evaluation.

\begin{table}[h]
\centering
\small
\begin{tabular}{c|cc}
\toprule[1.2pt]
Scenario & NMAE $\downarrow$ & Spearman-Rho $\uparrow$ \\
\midrule

Overall & 0.0130 & 0.9378 \\
\bottomrule[1.2pt]
\end{tabular}
\caption{Overall NMAE \& Spearman-Rho vs. Expert Annotations }
\label{tab:scenarioSpearmanScore}
\end{table}

\textbf{Performance on open-source dataset}

To validate the generality and robustness of our approach, extensive experiments were conducted on three widely used public jailbreak evaluation datasets—JBB, JailJudge, and Safe-RLHF—which also include samples beyond our 14 defined scenarios, enabling verification of performance on unseen edge cases and emerging risk types (results in Table~\ref{tab:t2}).

\begin{itemize}
    \item On JBB, our method achieves the highest F1 score (0.99), outperforming Llamaguard3 (0.98) and substantially exceeding StringMatching (0.86) and Beaver (0.61).
    
    \item On JailJudge, it sets a new SOTA with an F1 score of 0.8241, surpassing JailJudge (0.8089) and Qi2023 (0.8012).

    \item On Safe-RLHF, despite Beaver (specifically fine-tuned on this dataset) leading, our method ranks second with an F1 score of 0.83, outperforming JailJudge (0.81) and Qi2023 (0.79).
\end{itemize}

Overall, our method achieves SOTA on JBB and JailJudge, and strong performance on Safe-RLHF, demonstrating superior performance across diverse datasets.

\begin{table*}[h]
\centering
\small
\begin{tabular}{c|cccc|cccc|cccc}
\toprule[1.2pt]
Datasets & \multicolumn{4}{c|}{JBB} & \multicolumn{4}{c|}{JailJudge} & \multicolumn{4}{c}{Safe-RLHF} \\ 
\cmidrule(lr){2-5} \cmidrule(lr){6-9} \cmidrule(lr){10-13}
Indicators & acc & prec & rec & f1 & acc & prec & rec & f1 & acc & prec & rec & f1 \\
\midrule
StringMatch & 0.86  & 0.88  & 0.84   & 0.86 & 0.72 & 0.57 & 0.68 & 0.62 & 0.60 &\textbf{ 0.99} & 0.59 & 0.74 \\
llamaguard2  & 0.86  & 0.96  & 0.74   & 0.84 & 0.82 & 0.76 & 0.66 & 0.71 & 0.75 & 0.62 & 0.93 & 0.74 \\
llamaguard3  & \underline{0.98} & 0.95 & \textbf{1.0} & \underline{0.98}  & 0.83 & 0.72 & 0.81 & 0.77 & 0.72 & 0.53 & \textbf{0.96} & 0.68 \\
Beaver       & 0.72  & 0.96 & 0.45 & 0.61  & 0.80 & \textbf{0.80} & 0.55 & 0.65 & \textbf{0.90} & 0.88 & \underline{0.94} & \textbf{0.91} \\
Qi2023 & 0.93 & 0.88 & 0.99  & 0.93 & 0.83 & 0.68 & \textbf{0.98} & 0.80 & 0.80 & 0.69 & 0.94 & 0.79 \\
JailJudge    & 0.96 & \underline{0.99} & 0.93 & 0.96 & \underline{0.87} & \underline{0.80} & 0.82 & \underline{0.81} & 0.81 & 0.76 & 0.88 & 0.81 \\
Ours        & \textbf{0.99}  & \textbf{1.0} & \underline{0.99} & \textbf{0.99} & \textbf{0.87} & 0.76 & \underline{0.90} & \textbf{0.82} & \underline{0.81} &\underline{0.89}  & 0.77 & \underline{0.83} \\
\bottomrule[1.2pt]
\end{tabular}
\caption{Jailbreak evaluation performance on JBB, JailJudge, and Safe-RLHF datasets.}
\label{tab:t2}
\end{table*}

\subsection{Ablation Study}

We progressively remove SceneJailEval’s two key components while freezing the LLM backbone and prompts.  
Discarding scenario classification (DimsOnly) still yields high recall (91.8 \%) yet lowers F1 by 2.7 \%, indicating that uniform dimension scoring introduces false positives.  
Further removing dimension selection and reverting to vanilla heuristic rules (Vanilla) causes an additional 8.6 \% F1 drop (83.1 \%), confirming the necessity of scenario-adaptive evaluation.  
Overall, the full SceneJailEval achieves the best balance with 91.7 \% F1 (Table \ref{tab:ablation}).
\begin{table}[H]
\centering
\small
\begin{tabular}{c|cccc}
\toprule[1.2pt]
Method & accuracy & precision & recall & f1 \\
\midrule
Vanilla & 0.7676 & 0.8333 & 0.8296 & 0.8314 \\
SceneOnly & 0.7829 & 0.8444 & 0.8407 & 0.8425 \\
DimsOnly & 0.8440 & 0.8646 & 0.9181 & 0.8903 \\
Ours & \textbf{0.8900} & \textbf{0.8951} & \textbf{0.9398} & \textbf{0.9169} \\
\bottomrule[1.2pt]
\end{tabular}
\caption{Ablation study on SceneJailEval dataset.}
\label{tab:ablation}
\end{table}

\begin{table*}[t]
\centering
\small
\begin{tabular}{c|cc|cc|cc|cc|cc}
\toprule[1.2pt]
Model & \multicolumn{2}{c|}{GPT-4o} & \multicolumn{2}{c|}{Claude-3.5} & \multicolumn{2}{c|}{Gemini-2.5} & \multicolumn{2}{c|}{Llama-3-8b} & \multicolumn{2}{c}{Llama-2-7b} \\
\cmidrule(lr){2-3} \cmidrule(lr){4-5} \cmidrule(lr){6-7} \cmidrule(lr){8-9} \cmidrule(lr){10-11}
Indicators & ASR & $\overline{\text{Harm}}$  & ASR & $\overline{\text{Harm}}$  & ASR & $\overline{\text{Harm}}$  & ASR & $\overline{\text{Harm}}$  & ASR & $\overline{\text{Harm}}$  \\
\midrule

Violent Crime & 0\% & 0 & 0\% & 0 & 3.85\% & 0.13 & 3.85\% & 0.15 & 0\% & 0 \\
Non-violent Crime & 17.65\% & 0.50 & 2.61\% & 0.04 & 15.36\% & 0.47 & 0.65\% & 0.01 & 3.92\% & 0.09 \\
Sex-related Crime & 0\% & 0 & 0\% & 0 & 0\% & 0 & 0\% & 0 & 0\% & 0 \\
Child Sexual Exploit. & 10.71\% & 0.07 & 10.71\% & 0.08 & 0\% & 0 & 0\% & 0 & 0\% & 0 \\
False Info. \& Defamation & 6.72\% & 0.26 & 0\% & 0 & 4.48\% & 0.13 & 8.96\% & 0.23 & 2.24\% & 0.04 \\
Professional Advice & 7.27\% & 0.16 & 1.82\% & 0.02 & 1.82\% & 0.02 & 0\% & 0 & 7.27\% & 0.18 \\
Privacy Invasion & 3.64\% & 0.13 & 0\% & 0 & 2.73\% & 0.09 & 1.82\% & 0.04 & 3.64\% & 0.07 \\
IP Infringement & 10.94\% & 0.29 & 0.78\% & 0.02 & 9.38\% & 0.15 & 18.75\% & 0.34 & 15.62\% & 0.27 \\
WMDs & 3.88\% & 0.13 & 0.97\% & 0.02 & 3.88\% & 0.15 & 1.94\% & 0.08 & 0.97\% & 0.00 \\
Hate \& Discrimination & 1.06\% & 0.03 & 2.13\% & 0.04 & 0\% & 0 & 0\% & 0 & 0\% & 0 \\
Suicide \& Self-harm & 6.87\% & 0.21 & 0\% & 0 & 0\% & 0 & 0\% & 0 & 0\% & 0 \\
Sexual Content & 25.00\% & 0.63 & 0\% & 0 & 0\% & 0 & 0\% & 0 & 7.14\% & 0.10 \\
Pol. Agitation \& Election & 17.00\% & 0.63 & 0\% & 0 & 2.00\% & 0.07 & 3.00\% & 0.10 & 3.00\% & 0.04 \\
Regional Sens. Issues & 70.37\% & 2.76 & 8.33\% & 0.20 & 23.15\% & 0.74 & 29.63\% & 0.99 & 25.93\% & 0.87 \\
Overall & 15.06\% & 0.50 & 1.91\% & 0.03 & 9.11\% & 0.21 & 6.88\% & 0.14 & 5.89\% & 0.14 \\
\bottomrule[1.2pt]
\end{tabular}
\caption{Cross-scenario LLM security assessment via SceneJailEval.}
\label{tab:LLMs}
\end{table*}

\subsection{Comprehensive Security Evaluation of LLMs}

Using SceneJailEval, we evaluated mainstream LLMs (Gemini2.5Flash, GPT-4o, Claude 3.5, Llama) via joint ASR and Average Harm Score ($\overline{\text{Harm}}$) assessment.
 Results reveal Claude-3.5 as most robust (Avg Harm: 0.033) and GPT-4o as most vulnerable (0.502; Table~\ref{tab:LLMs}).  

Scenario-wise analysis uncovers a critical correlation: Regional Sensitive Issues consistently challenge all models, while Sex-related Crime elicits strong resilience—highlighting scenario-dependent security mechanisms.  

Notably, each model exhibits unique vulnerability profiles: GPT-4o (25\% ASR in Sexual Content), Claude-3.5 (10.71\% in Child Sexual Exploitation), Gemini-2.5-Flash (15.36\% in Non-violent Crime), and Llama variants (IP Infringement). These findings highlight the heterogeneous vulnerability landscape across models, informing robustness enhancements against diverse jailbreak threats.

\subsection{Case Study on Customized Scenario}\label{subsec:case-study}

To address diverse organizational compliance needs, our framework is engineered with exceptional extensibility, seamlessly supporting customized detection scenarios and evaluation dimensions. To validate this, we designed a ``Product Consultation" custom scenario (requiring models to avoid self-product derogation), where our Custom Generation Agent automatically generated the scenario category, detection dimension (Loyalty), and harm dimensions (Derogation, Specificity). Evaluating a 200-query custom dataset via SceneJailEval’s annotation protocol (Table~\ref{tab:exp4}) yielded stellar results, conclusively demonstrating the framework’s robust extensibility and precise evaluation in tailored scenarios—underscoring its practical versatility.

\begin{table}[htbp]
\centering
\small
\begin{tabular}{c|c|c|c|c|c}
\toprule[1pt]
acc & prec & rec & f1 & NMAE & Spearman-Rho \\
\midrule
1.0 & 1.0 & 1.0 & 1.0 & 0.037 & 0.841 \\
\bottomrule[1pt]
\end{tabular}
\caption{Jailbreak evaluation performance on the customized scenario.}
\label{tab:exp4}
\end{table}

\section{Conclusion}

SceneJailEval revolutionizes LLM jailbreak evaluation with a paradigm-shifting scenario-adaptive framework, eliminating ``one-size-fits-all" limitations and offering seamless extensibility for diverse needs. Complemented by a groundbreaking multi-scenario dataset—rich in variants and regional cases—it fills the critical gap in high-quality scenario-aware benchmarks. Boasting SOTA performance (0.917 F1 on our dataset, +6\% over prior; 0.995 F1 on JBB, +3\% over prior), it shatters accuracy bottlenecks in heterogeneous scenarios. These advances set a new standard for context-aware LLM security assessment, strengthening jailbreak defenses and accelerating trustworthy AI progress.

\section{Acknowledgments}
This work is supported by the National Natural Science Foundation of China under Grant Nos. 62572316 and 62302303, the Natural Science Foundation of Shanghai under Grant Nos. 25ZR1402279 and 23ZR1434000, and the Shanghai Pujiang Program under Grant No. 24PJD043.


\bibliography{aaai2026}

@article{liao2024amplegcg,
  title={Amplegcg: Learning a universal and transferable generative model of adversarial suffixes for jailbreaking both open and closed llms},
  author={Liao, Zeyi and Sun, Huan},
  journal={arXiv preprint arXiv:2404.07921},
  year={2024}
}

@article{abdin2024phi,
  title={Phi-4 technical report},
  author={Abdin, Marah and Aneja, Jyoti and Behl, Harkirat and Bubeck, S{\'e}bastien and Eldan, Ronen and Gunasekar, Suriya and Harrison, Michael and Hewett, Russell J and Javaheripi, Mojan and Kauffmann, Piero and others},
  journal={arXiv preprint arXiv:2412.08905},
  year={2024}
}

@article{saaty1980analytic,
  title={The analytic hierarchy process mcgraw hill, New York},
  author={Saaty, Thomas L},
  journal={Agricultural Economics Review},
  volume={70},
  number={804},
  pages={10--21236},
  year={1980}
}

@article{dalkey1963experimental,
  title={An experimental application of the Delphi method to the use of experts},
  author={Dalkey, Norman and Helmer, Olaf},
  journal={Management science},
  volume={9},
  number={3},
  pages={458--467},
  year={1963},
  publisher={INFORMS}
}

@inproceedings{zhang2024jailbreak,
  title={Jailbreak open-sourced large language models via enforced decoding},
  author={Zhang, Hangfan and Guo, Zhimeng and Zhu, Huaisheng and Cao, Bochuan and Lin, Lu and Jia, Jinyuan and Chen, Jinghui and Wu, Dinghao},
  booktitle={Proceedings of the 62nd Annual Meeting of the Association for Computational Linguistics (Volume 1: Long Papers)},
  pages={5475--5493},
  year={2024}
}

@article{yuan2024rigorllm,
  title={Rigorllm: Resilient guardrails for large language models against undesired content},
  author={Yuan, Zhuowen and Xiong, Zidi and Zeng, Yi and Yu, Ning and Jia, Ruoxi and Song, Dawn and Li, Bo},
  journal={arXiv preprint arXiv:2403.13031},
  year={2024}
}

@article{paulus2024advprompter,
  title={Advprompter: Fast adaptive adversarial prompting for llms},
  author={Paulus, Anselm and Zharmagambetov, Arman and Guo, Chuan and Amos, Brandon and Tian, Yuandong},
  journal={arXiv preprint arXiv:2404.16873},
  year={2024}
}

@article{huang2025guidedbench,
  title={Guidedbench: Equipping jailbreak evaluation with guidelines},
  author={Huang, Ruixuan and Wang, Xunguang and Li, Zongjie and Wu, Daoyuan and Wang, Shuai},
  journal={arXiv preprint arXiv:2502.16903},
  year={2025}
}

@article{mell2007common,
  title={Common vulnerability scoring system},
  author={Mell, Peter and Scarfone, Karen and Romanosky, Sasha},
  journal={IEEE Security \& Privacy},
  volume={4},
  number={6},
  pages={85--89},
  year={2007},
  publisher={IEEE}
}

@book{howard2003writing,
  title={Writing secure code},
  author={Howard, Michael and LeBlanc, David},
  year={2003},
  publisher={Pearson Education}
}

@inproceedings{shen2024anything,
  title={" do anything now": Characterizing and evaluating in-the-wild jailbreak prompts on large language models},
  author={Shen, Xinyue and Chen, Zeyuan and Backes, Michael and Shen, Yun and Zhang, Yang},
  booktitle={Proceedings of the 2024 on ACM SIGSAC Conference on Computer and Communications Security},
  pages={1671--1685},
  year={2024}
}

@article{fu2023gptscore,
  title={Gptscore: Evaluate as you desire},
  author={Fu, Jinlan and Ng, See-Kiong and Jiang, Zhengbao and Liu, Pengfei},
  journal={arXiv preprint arXiv:2302.04166},
  year={2023}
}

@inproceedings{chao2025jailbreaking,
  title={Jailbreaking black box large language models in twenty queries},
  author={Chao, Patrick and Robey, Alexander and Dobriban, Edgar and Hassani, Hamed and Pappas, George J and Wong, Eric},
  booktitle={2025 IEEE Conference on Secure and Trustworthy Machine Learning (SaTML)},
  pages={23--42},
  year={2025},
  organization={IEEE}
}

@article{mazeika2024harmbench,
  title={Harmbench: A standardized evaluation framework for automated red teaming and robust refusal},
  author={Mazeika, Mantas and Phan, Long and Yin, Xuwang and Zou, Andy and Wang, Zifan and Mu, Norman and Sakhaee, Elham and Li, Nathaniel and Basart, Steven and Li, Bo and others},
  journal={arXiv preprint arXiv:2402.04249},
  year={2024}
}

@article{zeng2024autodefense,
  title={Autodefense: Multi-agent llm defense against jailbreak attacks},
  author={Zeng, Yifan and Wu, Yiran and Zhang, Xiao and Wang, Huazheng and Wu, Qingyun},
  journal={arXiv preprint arXiv:2403.04783},
  year={2024}
}

@article{du2023analyzing,
  title={Analyzing the inherent response tendency of llms: Real-world instructions-driven jailbreak},
  author={Du, Yanrui and Zhao, Sendong and Ma, Ming and Chen, Yuhan and Qin, Bing},
  journal={arXiv preprint arXiv:2312.04127},
  year={2023}
}

@article{ding2023wolf,
  title={A wolf in sheep's clothing: Generalized nested jailbreak prompts can fool large language models easily},
  author={Ding, Peng and Kuang, Jun and Ma, Dan and Cao, Xuezhi and Xian, Yunsen and Chen, Jiajun and Huang, Shujian},
  journal={arXiv preprint arXiv:2311.08268},
  year={2023}
}

@article{gehman2020realtoxicityprompts,
  title={Realtoxicityprompts: Evaluating neural toxic degeneration in language models},
  author={Gehman, Samuel and Gururangan, Suchin and Sap, Maarten and Choi, Yejin and Smith, Noah A},
  journal={arXiv preprint arXiv:2009.11462},
  year={2020}
}

@article{ghosh2025ailuminate,
  title={Ailuminate: Introducing v1. 0 of the ai risk and reliability benchmark from mlcommons},
  author={Ghosh, Shaona and Frase, Heather and Williams, Adina and Luger, Sarah and R{\"o}ttger, Paul and Barez, Fazl and McGregor, Sean and Fricklas, Kenneth and Kumar, Mala and Bollacker, Kurt and others},
  journal={arXiv preprint arXiv:2503.05731},
  year={2025}
}

@article{cheng2024soft,
  title={Soft-label integration for robust toxicity classification},
  author={Cheng, Zelei and Wu, Xian and Yu, Jiahao and Han, Shuo and Cai, Xin-Qiang and Xing, Xinyu},
  journal={Advances in Neural Information Processing Systems},
  volume={37},
  pages={94776--94807},
  year={2024}
}

@article{yu2022hate,
  title={Hate speech and counter speech detection: Conversational context does matter},
  author={Yu, Xinchen and Blanco, Eduardo and Hong, Lingzi},
  journal={arXiv preprint arXiv:2206.06423},
  year={2022}
}

@misc{eu_artificial_intelligence_act,
  title = {Artificial Intelligence Act},
  author = {EU},
  year = {2024},
  note = {Retrieved from https://artificialintelligenceact.eu/}
}

@article{rauh2022characteristics,
  title={Characteristics of harmful text: Towards rigorous benchmarking of language models},
  author={Rauh, Maribeth and Mellor, John and Uesato, Jonathan and Huang, Po-Sen and Welbl, Johannes and Weidinger, Laura and Dathathri, Sumanth and Glaese, Amelia and Irving, Geoffrey and Gabriel, Iason and others},
  journal={Advances in Neural Information Processing Systems},
  volume={35},
  pages={24720--24739},
  year={2022}
}

@misc{nist_ai_risk_management_framework,
  title = {Artificial Intelligence Risk Management Framework},
  author = {NIST},
  year = {2023},
  note = {Retrieved from https://www.nist.gov/itl/ai-risk-management-framework}
}

@article{sutcliffe1998supporting,
  title={Supporting scenario-based requirements engineering},
  author={Sutcliffe, Alistair G and Maiden, Neil AM and Minocha, Shailey and Manuel, Darrel},
  journal={IEEE Transactions on software engineering},
  volume={24},
  number={12},
  pages={1072--1088},
  year={1998},
  publisher={IEEE}
}

@article{ryser1999scenario,
  title={A scenario-based approach to validating and testing software systems using statecharts},
  author={Ryser, Johannes and Glinz, Martin},
  year={1999},
  publisher={CNAM}
}

@inproceedings{nalic2020scenario,
  title={Scenario based testing of automated driving systems: A literature survey},
  author={Nalic, Demin and Mihalj, Tomislav and B{\"a}umler, Maximilian and Lehmann, Matthias and Eichberger, Arno and Bernsteiner, Stefan},
  booktitle={FISITA web Congress},
  volume={10},
  year={2020}
}

@article{sun2021scenario,
  title={Scenario-based test automation for highly automated vehicles: A review and paving the way for systematic safety assurance},
  author={Sun, Jian and Zhang, He and Zhou, Huajun and Yu, Rongjie and Tian, Ye},
  journal={IEEE transactions on intelligent transportation systems},
  volume={23},
  number={9},
  pages={14088--14103},
  year={2021},
  publisher={IEEE}
}

@article{lapid2023open,
  title={Open sesame! universal black box jailbreaking of large language models},
  author={Lapid, Raz and Langberg, Ron and Sipper, Moshe},
  journal={arXiv preprint arXiv:2309.01446},
  year={2023}
}

@article{liu2023autodan,
  title={Autodan: Generating stealthy jailbreak prompts on aligned large language models},
  author={Liu, Xiaogeng and Xu, Nan and Chen, Muhao and Xiao, Chaowei},
  journal={arXiv preprint arXiv:2310.04451},
  year={2023}
}

@article{zhang2024intention,
  title={Intention analysis makes llms a good jailbreak defender},
  author={Zhang, Yuqi and Ding, Liang and Zhang, Lefei and Tao, Dacheng},
  journal={arXiv preprint arXiv:2401.06561},
  year={2024}
}

@article{cai2024rethinking,
  title={Rethinking how to evaluate language model jailbreak},
  author={Cai, Hongyu and Arunasalam, Arjun and Lin, Leo Y and Bianchi, Antonio and Celik, Z Berkay},
  journal={arXiv preprint arXiv:2404.06407},
  year={2024}
}

@inproceedings{banerjee2025ethical,
  title={How (un) ethical are instruction-centric responses of llms? unveiling the vulnerabilities of safety guardrails to harmful queries},
  author={Banerjee, Somnath and Layek, Sayan and Hazra, Rima and Mukherjee, Animesh},
  booktitle={Proceedings of the International AAAI Conference on Web and Social Media},
  volume={19},
  pages={193--205},
  year={2025}
}

@article{liu2023robustness,
  title={Robustness Over Time: Understanding Adversarial Examples' Effectiveness on Longitudinal Versions of Large Language Models},
  author={Liu, Yugeng and Cong, Tianshuo and Zhao, Zhengyu and Backes, Michael and Shen, Yun and Zhang, Yang},
  journal={arXiv preprint arXiv:2308.07847},
  year={2023}
}

@article{zheng2023judging,
  title={Judging llm-as-a-judge with mt-bench and chatbot arena},
  author={Zheng, Lianmin and Chiang, Wei-Lin and Sheng, Ying and Zhuang, Siyuan and Wu, Zhanghao and Zhuang, Yonghao and Lin, Zi and Li, Zhuohan and Li, Dacheng and Xing, Eric and others},
  journal={Advances in neural information processing systems},
  volume={36},
  pages={46595--46623},
  year={2023}
}

@article{xiao2024distract,
  title={Distract large language models for automatic jailbreak attack},
  author={Xiao, Zeguan and Yang, Yan and Chen, Guanhua and Chen, Yun},
  journal={arXiv preprint arXiv:2403.08424},
  year={2024}
}

@inproceedings{liu2024making,
  title={Making them ask and answer: Jailbreaking large language models in few queries via disguise and reconstruction},
  author={Liu, Tong and Zhang, Yingjie and Zhao, Zhe and Dong, Yinpeng and Meng, Guozhu and Chen, Kai},
  booktitle={33rd USENIX Security Symposium (USENIX Security 24)},
  pages={4711--4728},
  year={2024}
}

@article{qiu2023latent,
  title={Latent jailbreak: A benchmark for evaluating text safety and output robustness of large language models},
  author={Qiu, Huachuan and Zhang, Shuai and Li, Anqi and He, Hongliang and Lan, Zhenzhong},
  journal={arXiv preprint arXiv:2307.08487},
  year={2023}
}

@article{huang2023catastrophic,
  title={Catastrophic jailbreak of open-source llms via exploiting generation},
  author={Huang, Yangsibo and Gupta, Samyak and Xia, Mengzhou and Li, Kai and Chen, Danqi},
  journal={arXiv preprint arXiv:2310.06987},
  year={2023}
}

@article{zou2023universal,
  title={Universal and transferable adversarial attacks on aligned language models},
  author={Zou, Andy and Wang, Zifan and Carlini, Nicholas and Nasr, Milad and Kolter, J Zico and Fredrikson, Matt},
  journal={arXiv preprint arXiv:2307.15043},
  year={2023}
}

@article{shu2025attackeval,
  title={Attackeval: How to evaluate the effectiveness of jailbreak attacking on large language models},
  author={Shu, Dong and Zhang, Chong and Jin, Mingyu and Zhou, Zihao and Li, Lingyao},
  journal={ACM SIGKDD Explorations Newsletter},
  volume={27},
  number={1},
  pages={10--19},
  year={2025},
  publisher={ACM New York, NY, USA}
}

@article{souly2024strongreject,
  title={A strongreject for empty jailbreaks},
  author={Souly, Alexandra and Lu, Qingyuan and Bowen, Dillon and Trinh, Tu and Hsieh, Elvis and Pandey, Sana and Abbeel, Pieter and Svegliato, Justin and Emmons, Scott and Watkins, Olivia and others},
  journal={Advances in Neural Information Processing Systems},
  volume={37},
  pages={125416--125440},
  year={2024}
}

@inproceedings{Beaver,
 author = {Ji, Jiaming and Liu, Mickel and Dai, Josef and Pan, Xuehai and Zhang, Chi and Bian, Ce and Chen, Boyuan and Sun, Ruiyang and Wang, Yizhou and Yang, Yaodong},
 booktitle = {Advances in Neural Information Processing Systems},
 editor = {A. Oh and T. Naumann and A. Globerson and K. Saenko and M. Hardt and S. Levine},
 pages = {24678--24704},
 publisher = {Curran Associates, Inc.},
 title = {BeaverTails: Towards Improved Safety Alignment of LLM via a Human-Preference Dataset},
 url = {https://proceedings.neurips.cc/paper_files/paper/2023/file/4dbb61cb68671edc4ca3712d70083b9f-Paper-Datasets_and_Benchmarks.pdf},
 volume = {36},
 year = {2023}
}

@misc{llamaguard2,
      title={Llama Guard: LLM-based Input-Output Safeguard for Human-AI Conversations}, 
      author={Hakan Inan and Kartikeya Upasani and Jianfeng Chi and Rashi Rungta and Krithika Iyer and Yuning Mao and Michael Tontchev and Qing Hu and Brian Fuller and Davide Testuggine and Madian Khabsa},
      year={2023},
      eprint={2312.06674},
      archivePrefix={arXiv},
      primaryClass={cs.CL},
      url={https://arxiv.org/abs/2312.06674}, 
}

@misc{llamaguard3,
      title={Llama Guard 3 Vision: Safeguarding Human-AI Image Understanding Conversations}, 
      author={Jianfeng Chi and Ujjwal Karn and Hongyuan Zhan and Eric Smith and Javier Rando and Yiming Zhang and Kate Plawiak and Zacharie Delpierre Coudert and Kartikeya Upasani and Mahesh Pasupuleti},
      year={2024},
      eprint={2411.10414},
      archivePrefix={arXiv},
      primaryClass={cs.CV},
      url={https://arxiv.org/abs/2411.10414}, 
}

@article{qi2023,
title = "Visual Adversarial Examples Jailbreak Aligned Large Language Models",
author = "Xiangyu Qi and Kaixuan Huang and Ashwinee Panda and Peter Henderson and Mengdi Wang and Prateek Mittal",
note = "Publisher Copyright: Copyright {\textcopyright} 2024, Association for the Advancement of Artificial Intelligence (www.aaai.org). All rights reserved.; 38th AAAI Conference on Artificial Intelligence, AAAI 2024 ; Conference date: 20-02-2024 Through 27-02-2024",
year = "2024",
month = mar,
day = "25",
doi = "10.1609/aaai.v38i19.30150",
language = "English (US)",
volume = "38",
pages = "21527--21536",
journal = "Proceedings of the AAAI Conference on Artificial Intelligence",
issn = "2159-5399",
publisher = "Association for the Advancement of Artificial Intelligence",
number = "19",
}

@misc{jailjudge,
      title={JAILJUDGE: A Comprehensive Jailbreak Judge Benchmark with Multi-Agent Enhanced Explanation Evaluation Framework}, 
      author={Fan Liu and Yue Feng and Zhao Xu and Lixin Su and Xinyu Ma and Dawei Yin and Hao Liu},
      year={2024},
      eprint={2410.12855},
      archivePrefix={arXiv},
      primaryClass={cs.CL},
      url={https://arxiv.org/abs/2410.12855}, 
}

@article{chao2024jailbreakbench,
title={Jailbreakbench: An open robustness benchmark for jailbreaking large language models},
author={Chao, Patrick and Debenedetti, Edoardo and Robey, Alexander and Andriushchenko, Maksym and Croce, Francesco and Sehwag, Vikash and Dobriban, Edgar and Flammarion, Nicolas and Pappas, George J and Tramer, Florian and others},
journal={Advances in Neural Information Processing Systems},
volume={37},
pages={55005--55029},
year={2024}
}

@inproceedings{safe-rlhf,
  title={Safe RLHF: Safe Reinforcement Learning from Human Feedback},
  author={Josef Dai and Xuehai Pan and Ruiyang Sun and Jiaming Ji and Xinbo Xu and Mickel Liu and Yizhou Wang and Yaodong Yang},
  booktitle={The Twelfth International Conference on Learning Representations},
  year={2024},
  url={https://openreview.net/forum?id=TyFrPOKYXw}
}

\end{document}